\documentclass{article}

\PassOptionsToPackage{numbers, compress}{natbib}
\usepackage[main, final]{neurips_2026}

\usepackage[utf8]{inputenc}
\usepackage[T1]{fontenc}
\usepackage{url}
\usepackage{booktabs}
\usepackage{amsfonts}
\usepackage{amsmath}
\usepackage{amssymb}
\usepackage{nicefrac}
\usepackage{microtype}
\usepackage{xcolor}
\usepackage{graphicx}
\usepackage{subcaption}
\usepackage{algorithm}
\usepackage{algorithmic}
\usepackage{multirow}

\title{Anchor3DGS: Feed-Forward 3D Gaussian Splatting with Compact Anchor-Based Representation}

\author{%
  Sheng Ye$^{1,2,*}$
  , Jiangke Lin$^{2}$, Fudong Wang$^{2}$, Yi Yuan$^{2,\dag}$, Zhen-Hui Dong$^{1}$, Yong-Jin Liu$^{1,\ddag}$ \\
  $^{1}$Tsinghua University \\
  $^{2}$Ant Group, Inclusion AI \\
  \texttt{\{yec22, dzh23\}@mails.tsinghua.edu.cn}, \texttt{liuyongjin@tsinghua.edu.cn} \\
  \texttt{\{linjiangke.ljk, wangfudong.wfd, yy464236\}@antgroup.com}
}

\begin{document}

\maketitle

\begingroup
  \renewcommand{\thefootnote}{*}
  \footnotetext{Work done during an internship at Ant Group.}
  
  \renewcommand{\thefootnote}{\dag}
  \footnotetext{Project lead.}
  
  \renewcommand{\thefootnote}{\ddag}
  \footnotetext{Corresponding author.}
\endgroup

\begin{abstract}
Recent feed-forward 3D Gaussian Splatting (3DGS) methods have enabled effective multi-view 3D reconstruction by predicting Gaussian primitives in a single forward pass. However, existing approaches predominantly adopt a pixel-aligned paradigm that predicts one Gaussian per input pixel, coupling the Gaussian count to the input resolution and number of views. This also leads to redundant primitives in textureless regions and insufficient coverage in geometrically complex areas.
We present \textbf{Anchor3DGS}, a pose-free feed-forward 3DGS framework that replaces pixel-aligned prediction with a compact anchor-based representation. Guided by information entropy, our method places a set of 3D anchors in the scene, concentrating them in regions of higher visual complexity while maintaining broad spatial coverage. Each anchor performs occlusion-aware aggregation of multi-view image features, exchanges information with other anchors, and is decoded into a set of local Gaussians. A lightweight feature enhancement module further refines the coarse RGB output from Gaussian splatting and improves rendering fidelity. Extensive experiments on diverse benchmark datasets demonstrate that our \textbf{Anchor3DGS} achieves state-of-the-art novel view synthesis quality while only using an order of magnitude fewer Gaussian primitives.
\end{abstract}

\section{Introduction}

Reconstructing 3D scenes and synthesizing novel views from multiple 2D captures is a fundamental problem in computer vision and graphics, with broad applications in robotics, autonomous driving, gaming, and virtual reality.
The pioneering NeRF~\citep{nerf} represents scenes as an implicit radiance field optimized via volume rendering, while subsequent 3DGS~\citep{3DGS} employs differentiable rasterization of anisotropic 3D Gaussians to achieve real-time, high-fidelity rendering.
Despite its success, standard 3DGS requires a time-consuming per-scene optimization process. To address this limitation, a growing line of work has explored feed-forward 3DGS reconstruction~\citep{pixelsplat, mvsplat, depthsplat, mvsplat360, freesplat}, which trains generalizable models on large-scale datasets to directly predict 3D Gaussians from input views in a single forward pass. Recent advances have further eliminated the dependency on pre-computed camera poses~\citep{noposplat, anysplat, instantsplat}, enabling truly end-to-end reconstruction from uncalibrated images.

\begin{figure}[t]
    \centering
    \includegraphics[width=0.99\linewidth]{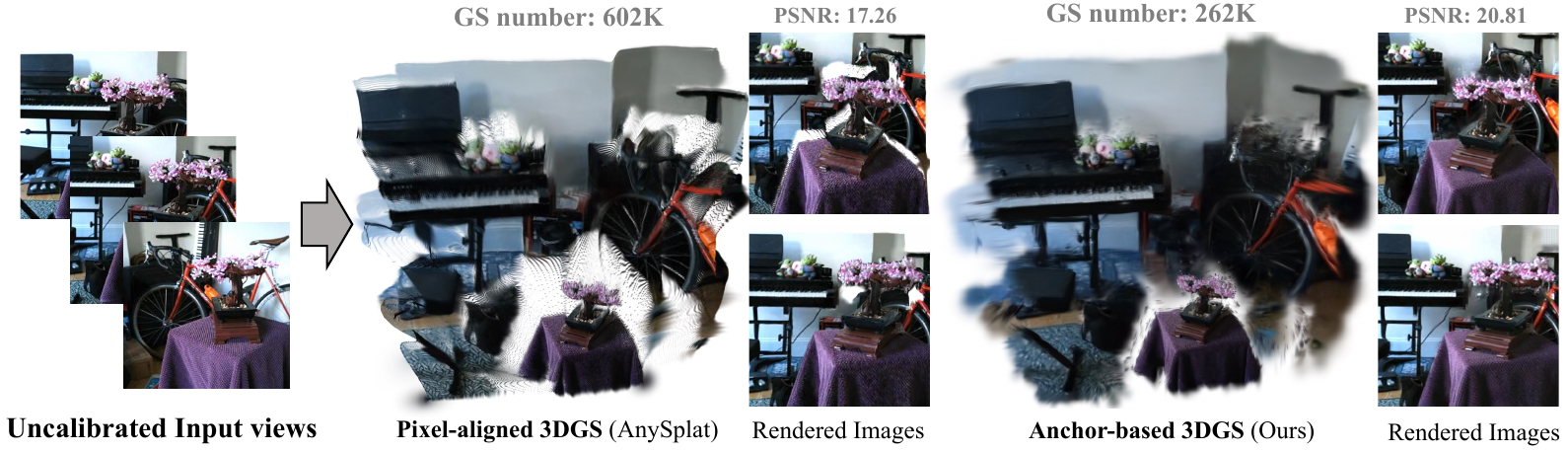}
    \caption{\textbf{Anchor3DGS} reconstructs 3D scenes from uncalibrated multi-view images. Our anchor-based representation achieves better rendering quality with fewer Gaussians, and avoids the empty holes that pixel-aligned methods (\textit{e.g.,} AnySplat~\citep{anysplat}) commonly leave in poorly observed regions.}
    \label{fig:teaser}
\end{figure}

However, current feed-forward approaches adopt a \textit{pixel-aligned} paradigm in which one Gaussian is predicted for each input pixel. While conceptually straightforward, this design suffers from three key limitations: (1) \textbf{Redundancy}---textureless or homogeneous regions consume as many Gaussians as detail-rich areas, yielding redundant primitives and wasted capacity; (2) \textbf{Insufficient coverage}---poorly observed regions cannot be faithfully reconstructed, since each Gaussian is predicted independently from a single pixel, without leveraging contextual information from well-observed neighbors; and (3) \textbf{Input dependency}---the total Gaussian count is strictly tied to the input resolution and view count rather than to the scene itself, preventing efficient reconstruction.

In this work, we present \textbf{Anchor3DGS}, a feed-forward framework that takes uncalibrated multi-view images as input and outputs 3D Gaussians of the scene (see Fig.~\ref{fig:teaser}). Our key insight is that a scene's 3D structure has no intrinsic connection to the number of input views or pixels that happen to observe it. We therefore propose to lift Gaussian allocation into 3D space: we represent the scene with a set of anchors whose \textit{spatial distribution} is guided by scene content, while the \textit{total number of anchors} is a hyperparameter, independent of image resolution and view count.

Our framework consists of several key components: \textit{(i) Entropy-guided anchor sampling.} Our 3D anchor sampling strategy simultaneously ensures broad spatial coverage of the scene and adaptively allocates more anchors to regions exhibiting higher visual complexity, as identified by local information entropy. Entropy automatically guides 3D anchors to where they are needed most: few anchors suffice for low-complexity regions (\textit{e.g.,} white walls or sky), while many anchors concentrate on high-complexity regions (\textit{e.g.,} intricate structures or rich textures).
\textit{(ii) Multi-view anchor aggregation and local decoding.} Each 3D anchor aggregates visual features from all input views under occlusion-aware weighting, and is decoded into a small set of neighboring Gaussians that represent its local 3D region. Anchors also exchange information with each other, allowing those in poorly observed regions to leverage the context from well-observed neighbors and still produce plausible appearances. In contrast, pixel-aligned methods often leave visible holes in such regions.
\textit{(iii) Feature enhancement.} Beyond standard attributes, each Gaussian is also assigned a feature vector. These features are later splatted and decoded by a lightweight network, which compensates for the high-frequency details missing from the coarse splatted RGB image.

Extensive experiments across benchmark datasets validate the effectiveness of our method. In sparse-view settings (3 and 6 input views), \textbf{Anchor3DGS} significantly outperforms existing state-of-the-art baselines. In dense-view settings ($32$ input views), \textbf{Anchor3DGS} achieves comparable rendering quality while requiring an order of magnitude fewer Gaussians. This demonstrates that a compact anchor-based representation can match the fidelity of pixel-aligned approaches with greater efficiency.
Our contributions can be summarized as follows:
\begin{itemize}
    \item We propose \textbf{Anchor3DGS}, a novel anchor-based feed-forward 3DGS framework that replaces the pixel-aligned paradigm with a compact anchor-based representation, decoupling the Gaussian count from input image resolution and the number of input views.
    \item We introduce three key designs—\textit{entropy-guided anchor sampling}, \textit{occlusion-aware anchor aggregation and local decoding}, and \textit{feature enhancement}—which together enable high-fidelity novel view synthesis with fine details.
    \item We achieve state-of-the-art performance on multiple benchmark datasets with substantially fewer Gaussians than existing methods, demonstrating that our method yields a more compact yet equally effective representation.
\end{itemize}

\section{Related Work}

\paragraph{Per-Scene Novel View Synthesis.}
Neural Radiance Fields (NeRF)~\citep{nerf} pioneered novel view synthesis by representing scenes as continuous density and radiance fields, and optimizing them via differentiable volume rendering. Despite subsequent improvements~\citep{ref-nerf, scaffold-gs, mipnerf360, instantngp, nerf++, frnerf}, the expensive ray sampling in volume rendering still prevents real-time performance.
3D Gaussian Splatting (3DGS)~\citep{3DGS, 3dgs-advances} instead represents scenes as explicit anisotropic 3D Gaussians and exploits differentiable tile-based rasterization, achieving real-time speed with superior quality.
However, both NeRF- and 3DGS-based methods require accurate camera poses, typically from a costly Structure-from-Motion (SfM) pipeline such as COLMAP~\citep{colmap}. To remove this dependency, several works~\citep{nope-nerf, cf3dgs, longsplat, robustness} attempt to jointly optimize the camera poses and the scene representation. 
Nevertheless, all these methods still rely on tedious per-scene optimization. In contrast, our \textbf{Anchor3DGS} directly predicts camera poses and 3D Gaussians in a feed-forward pass within seconds.

\paragraph{Feed-Forward Novel View Synthesis.}
Feed-forward novel view synthesis avoids per-scene optimization by learning generalizable priors from large-scale datasets. Early works~\citep{pixelnerf, ibrnet, mvsnerf} condition the NeRF representation on image features aggregated from input views, enabling cross-scene generalization without per-scene retraining.
Following the success of 3DGS, subsequent feed-forward works~\citep{splatter-image, pixelsplat, mvsplat, mvsplat360, depthsplat, freesplat, volsplat, gs-lrm} replace NeRF with explicit Gaussian primitives for higher rendering quality and faster speed. As representative examples, pixelSplat~\citep{pixelsplat} employs epipolar attention to infer a dense depth probability distribution for predicting 3D Gaussians from image pairs, whereas MVSplat~\citep{mvsplat} incorporates multi-view stereo principles by constructing explicit cost volumes for more accurate geometric reconstruction.
A recent line of pose-free feed-forward methods~\citep{noposplat, flare, yonosplat, anysplat} further eliminates the reliance on known camera parameters. By leveraging robust geometric priors from 3D foundation models~\citep{dust3r, vggt}, FLARE~\citep{flare} and AnySplat~\citep{anysplat} can simultaneously infer the camera poses of input images and the matching geometry-aligned 3D Gaussians.
Despite these advances, all aforementioned feed-forward 3DGS methods typically predict Gaussians in a dense, pixel-aligned manner, resulting in massive redundant Gaussians. Although AnySplat~\citep{anysplat} applies post-hoc voxelization for Gaussian reduction, this fails to address the root cause, and the remaining Gaussians are still far from compact. Instead, we decouple Gaussian allocation from the pixel-aligned paradigm and model the scene using a compact set of 3D anchors. A few concurrent works~\citep{c3g, tokengs} also attempt to explore compact representations using learnable tokens, whereas we explicitly leverage 3D scene structure and visual complexity to guide Gaussian allocation.

\section{Method}
The key idea of our method is to represent the scene with a compact set of 3D anchors, rather than being strictly bound to the input image pixels. Each anchor represents a local 3D region: it is placed on a geometry-aware scene scaffold, gathers information from multiple views, exchanges information with other anchors, and is finally decoded into a set of local Gaussians. In this way, the scene representation is organized directly in 3D space, with its size controlled by the anchor budget.

\begin{figure}[t]
    \centering
    \includegraphics[width=0.97\linewidth]{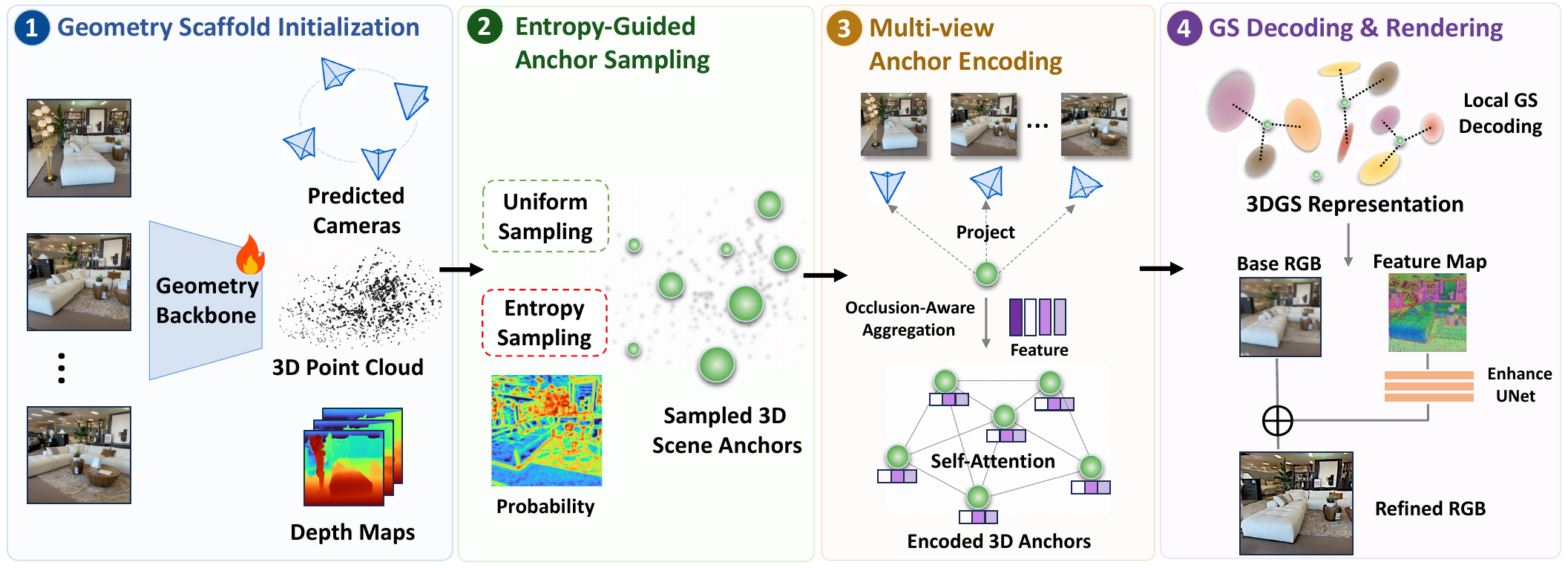}
    \caption{\textbf{Overview of Anchor3DGS.} Given uncalibrated multi-view images, \textit{(1)} a VGGT-based~\citep{vggt} geometry backbone first predicts the scene's geometric scaffold. \textit{(2)} A fixed budget of 3D anchors is then sampled from the scaffold, combining \textit{uniform} sampling with \textit{entropy-guided} sampling. \textit{(3)} Each anchor is projected back into all input views to gather features via occlusion-aware aggregation, and then interacts with other anchors through self-attention. \textit{(4)} Finally, each anchor is decoded into a set of local Gaussians, which are rasterized and refined to yield the final RGB output.}
    \label{fig:pipeline}
\end{figure}

Concretely, given a set of uncalibrated input images $\mathcal{I}=\{I_v\}_{v=1}^{V}$, we first build a multi-view geometric scaffold using a VGGT-based pose-free geometry backbone~\citep{vggt}, which predicts camera parameters $\mathcal{C}=\{C_v\}_{v=1}^{V}$ and per-view depth maps $\mathcal{D}=\{D_v\}_{v=1}^{V}$. By unprojecting the predicted depth maps with the predicted cameras, we obtain scene points $\mathcal{P}$ in 3D space, from which we sample a fixed number of anchors guided by local image entropy. Each anchor is then encoded into a latent token $\mathbf{z}_n$ through occlusion-aware multi-view feature aggregation and cross-anchor interaction. Finally, each token is decoded into $K$ local Gaussian primitives, yielding the final Gaussian set $\mathcal{G}=\{g_m\}_{m=1}^{NK}$. Fig.~\ref{fig:pipeline} illustrates the overall pipeline of \textbf{Anchor3DGS}.

\subsection{Pose-Free Geometry Initialization}
\label{sec:geometry}
Before constructing anchors, we first build a geometry-aware 3D scaffold of the scene. The role of this stage is to lift the uncalibrated input images into a consistent geometric space on which anchors can be placed and reasoned about. To this end, we employ a VGGT-based~\citep{vggt} backbone model as a strong multi-view geometry prior.

Given input images $\mathcal{I}=\{I_v\}_{v=1}^{V}$ with $I_v \in \mathbb{R}^{H \times W \times 3}$, the geometry backbone predicts for each view a camera extrinsic $E_v \in \mathbb{R}^{3 \times 4}$, a camera intrinsic $K_v \in \mathbb{R}^{3 \times 3}$, and a depth map $D_v \in \mathbb{R}^{H \times W}$. We denote the overall predicted camera parameters by $\mathcal{C}=\{C_v\}_{v=1}^{V}$ with $C_v=(E_v, K_v)$. Using $\mathcal{C}$, we unproject each depth map $D_v$ into a per-view 3D point map $\mathbf{P}_v \in \mathbb{R}^{H \times W \times 3}$, and denote the resulting set of 3D scene points by $\mathcal{P}=\{\mathbf{P}_v\}_{v=1}^{V}$.

This geometric scaffold serves two purposes in our method. First, the scene points in $\mathcal{P}$ provide a geometry-aware 3D candidate pool for anchor sampling. Second, the predicted camera parameters $\mathcal{C}$ allow every anchor to be projected back into the input views, which is essential for subsequent multi-view feature aggregation and occlusion reasoning. Based on this geometric scaffold, our final representation is built as a fixed-size set of 3D anchors with features in scene space.

\subsection{Entropy-Guided Anchor Sampling}
\label{sec:anchor}
\begin{figure}[t]
    \centering
    \includegraphics[width=0.98\linewidth]{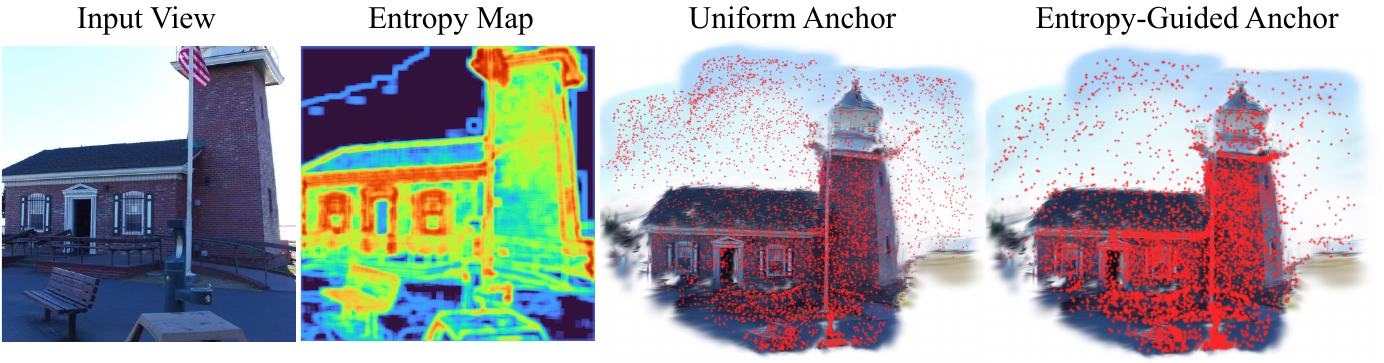}
    \caption{Uniform sampling ensures broad spatial coverage but wastes anchors (red dots) on uninformative regions (\textit{e.g.,} textureless sky). In contrast, our entropy-guided sampling allocates more anchors to visually complex regions while still maintaining coverage of the scene.}
    \label{fig:anchor}
\end{figure}

An effective allocation should both broadly cover the scene and place more anchors in visually complex regions where finer local representation is needed. To capture such complexity, we compute a Shannon entropy map~\citep{shannon, sparsesplat} for each input image. We first convert the RGB image to grayscale, quantize the grayscale values into bins, and compute a local histogram within a sliding window around each pixel. The entropy at image location $\mathbf{u}$ is then
\begin{equation}
E(\mathbf{u}) = - \sum_{b=1}^{B} p_b(\mathbf{u}) \log p_b(\mathbf{u}),
\end{equation}
where $p_b(\mathbf{u})$ is the fraction of pixels in the local window around $\mathbf{u}$ whose quantized grayscale values fall into the $b$-th bin. Regions with simple appearance, such as smooth walls or sky, tend to have low entropy, while textured structures and local boundaries tend to have high entropy. We use this quantity as a direct and training-free measure of local visual complexity (shown in Fig.~\ref{fig:anchor}).

Given the scene points $\mathcal{P}=\{\mathbf{P}_v\}_{v=1}^{V}$, we construct the anchor set using two complementary steps. We first perform coarse grid subsampling over the 3D points $\mathcal{P}$ to obtain anchors that broadly cover the scene. We then sample anchors from the full set of scene points with probabilities proportional to the entropy values at their associated image locations, so that more anchors are allocated to visually complex regions. Together, these two subsets form the final anchor set $\mathcal{A}$:
\begin{equation}
\mathcal{A} = \mathcal{A}_{\text{uni}} \cup \mathcal{A}_{\text{ent}} = \{\mathbf{a}_n\}_{n=1}^{N},\quad\mathbf{a}_n \in \mathbb{R}^{3},
\end{equation}
where $N$ is a fixed anchor budget. In this way, scene points provide anchor location candidates in 3D, while the final number of anchors is independent of the input resolution and the number of input views.
In addition, for each anchor $\mathbf{a}_n$, we estimate a local spacing value $d_n$ by averaging the distances to its $k$ nearest neighboring anchors:
\begin{equation}
d_n = \frac{1}{k}\sum_{j \in \mathcal{N}_k(n)} \|\mathbf{a}_n - \mathbf{a}_j\|_2.
\end{equation}
This spacing value provides a geometric density cue that will later be used to modulate the spatial extent of local Gaussian decoding.

\subsection{Occlusion-Aware Multi-View Anchor Aggregation}
With anchors allocated in scene space, the next step is to encode them for local Gaussian decoding. This differs from pixel-aligned Gaussian prediction, where each primitive corresponds to a single pixel observation from a specific view. In \textbf{Anchor3DGS}, each anchor represents a local 3D region in the scene, and is therefore encoded by aggregating information from multiple views.

For each anchor $\mathbf{a}_n$, we first compute an initial anchor embedding that captures both anchor position and local geometric scale information:
\begin{equation}
\mathbf{q}_n^{\text{geo}} = \mathrm{MLP}(\phi(\mathbf{a}_n, \log d_n)),
\end{equation}
where $\phi(\cdot)$ denotes Fourier positional encoding~\citep{nerf} applied to the 3D anchor coordinates and the logarithm of its local spacing $d_n$.
Using the predicted camera parameters $\mathcal{C}$, we project each anchor $\mathbf{a}_n$ onto every input view and bilinearly sample the feature map $\mathbf{F}_v$ (extracted by the VGGT-based backbone in Sec.~\ref{sec:geometry}) at the projected location, yielding a per-view anchor feature $\mathbf{f}_{n,v} \in \mathbb{R}^{C}$. For each sampled feature $\mathbf{f}_{n,v}$, we also estimate an occlusion weight $w^{\text{occ}}_{n,v} \in (0, 1)$, which attenuates the contribution of occluded views.
Specifically, we calculate $w^{\text{occ}}_{n,v}$ by comparing the camera-space depth $z_{n,v}$ of the anchor with the depth value $\tilde{z}_{n,v}$ sampled from $D_v$ (provided by the geometry backbone in Sec.~\ref{sec:geometry}) at the same projected location. If the anchor lies behind the visible surface, then $z_{n,v}$ will be noticeably larger than $\tilde{z}_{n,v}$, indicating likely occlusion; otherwise, the anchor is visible and its sampled feature can be trusted. We formalize this intuition as follows:
\begin{equation}
\Delta_{n,v} = \frac{z_{n,v} - \tilde{z}_{n,v}}{|\tilde{z}_{n,v}| + \epsilon}, \quad w^{\text{occ}}_{n,v} = \sigma(-\tau \Delta_{n,v}),
\end{equation}
where $\sigma(\cdot)$ is the sigmoid function and $\tau$ controls the sharpness to depth discrepancies.
We then aggregate the per-view features $\mathbf{f}_{n,v}$ biased by the occlusion weights:
\begin{equation}
\alpha_{n,v} = \mathrm{softmax}_v\big(\mathrm{MLP}([\mathbf{q}_n^{\text{geo}}, \mathbf{f}_{n,v}]) + \log w^{\text{occ}}_{n,v}\big), \quad \mathbf{f}_n^{\text{agg}} = \sum_{v=1}^{V} \alpha_{n,v} \, \mathbf{f}_{n,v}.
\end{equation}
The $\mathrm{MLP}$ scores each per-view feature's relevance to the anchor, while the additive $\log w^{\text{occ}}_{n,v}$ term suppresses occluded features. Together, they yield a weighting that jointly considers feature informativeness and geometric visibility.
Finally, the geometry-aware embedding $\mathbf{q}_n^{\text{geo}}$ and the aggregated feature $\mathbf{f}_n^{\text{agg}}$ are concatenated, and refined jointly across all $N$ anchors via \textit{self-attention} blocks: $\{\mathbf{z}_n \in \mathbb{R}^D\}_{n=1}^{N} = \mathrm{SelfAttn}(\{[\mathbf{q}_n^{\text{geo}}, \mathbf{f}_n^{\text{agg}}]\}_{n=1}^{N})$. This cross-anchor interaction allows anchors in weakly observed regions to draw context from better-observed neighbors, producing more coherent anchor embeddings $\{\mathbf{z}_n\}$ for subsequent Gaussian decoding.

\subsection{Anchor-to-Gaussian Decoding and Rendering}
After occlusion-aware aggregation, each anchor $\mathbf{a}_n$ is associated with an embedding $\mathbf{z}_n$. We then decode these embeddings into the final 3D Gaussians. Our goal is to keep the overall representation compact while still allowing each anchor to model a sufficiently rich local 3D region in both geometry and appearance. To this end, we decode a set of local Gaussians around each anchor.

Specifically, we introduce a set of $K$ learnable slot embeddings $\{\mathbf{s}_k\}_{k=1}^{K}$ (one per decoded Gaussian) shared across all anchors. For each anchor embedding $\mathbf{z}_n$, we combine it with each slot embedding $\mathbf{s}_k$ and decode both a local Gaussian offset $\Delta \mathbf{x}_{n,k}$ and a set of Gaussian attributes $\boldsymbol{\theta}_{n,k}$:
\begin{equation}
\Delta \mathbf{x}_{n,k} = \mathrm{MLP}_{\text{off}}([\mathbf{z}_n, \mathbf{s}_k]), \quad
\mathbf{x}_{n,k} = \mathbf{a}_n + d_n \, \Delta \mathbf{x}_{n,k}, \quad
\boldsymbol{\theta}_{n,k} = \mathrm{MLP}_{\text{gs}}([\mathbf{z}_n, \mathbf{s}_k]).
\end{equation}
Thus, the final center $\mathbf{x}_{n,k}$ of each decoded Gaussian is obtained by adding to the anchor $\mathbf{a}_n$ a local offset $\Delta \mathbf{x}_{n,k}$ scaled by the anchor's local spacing $d_n$ (discussed in Sec.~\ref{sec:anchor}), allowing anchors in sparse regions to decode Gaussians over a wider spatial extent. The decoded Gaussian attributes include: $\boldsymbol{\theta}_{n,k} = (o_{n,k}, \Sigma_{n,k}, \mathbf{c}_{n,k}, \mathbf{h}_{n,k})$, where $o$ is the opacity, $\Sigma$ is the covariance, $\mathbf{c}$ is the base RGB, and $\mathbf{h}$ is a feature vector that encodes high-frequency details for subsequent refinement. For simplicity, we re-index the resulting $NK$ Gaussians by a single index $i$ in the following.

To render a target view, we rasterize the predicted Gaussians using the differentiable pipeline of 3DGS~\citep{3DGS}. Each 3D Gaussian is first projected to 2D plane with mean $\boldsymbol{\mu}_i$ and covariance $\Sigma_i'$. The opacity contribution of Gaussian $i$ at pixel $\mathbf{u}$ is then modulated by:
\begin{equation}
\alpha_i(\mathbf{u}) = o_i \exp\!\Big(-\tfrac{1}{2}(\mathbf{u}-\boldsymbol{\mu}_i)^\top {\Sigma_i'}^{-1}(\mathbf{u}-\boldsymbol{\mu}_i)\Big).
\end{equation}
Sorting all Gaussians by depth and accumulating from front to back yields the transmittance $T_i(\mathbf{u}) = \prod_{j<i}(1-\alpha_j(\mathbf{u}))$. Together, $T_i$ and $\alpha_i$ serve as the alpha-compositing weights that, combined with the per-Gaussian color $\mathbf{c}_i$, give the standard 3DGS RGB rendering. In our setting, since each Gaussian additionally carries a feature vector $\mathbf{h}_i$, we also feed it through the same rasterizer to obtain a feature map $F_{\text{splat}}$ that is aligned with the rendered base image $I_{\text{base}}$:
\begin{equation}
I_{\text{base}}(\mathbf{u}) = \sum_i T_i(\mathbf{u})\,\alpha_i(\mathbf{u})\,\mathbf{c}_i, \quad
F_{\text{splat}}(\mathbf{u}) = \sum_i T_i(\mathbf{u})\,\alpha_i(\mathbf{u})\,\mathbf{h}_i.
\end{equation}
The two outputs are produced by the same compositing weights and therefore share identical geometric structures. The coarse image $I_{\text{base}}$ provides geometry-consistent base color signals, while the feature map $F_{\text{splat}}$ provides higher-dimensional appearance cues.
Finally, a lightweight UNet-based enhancement network takes $F_{\text{splat}}$ as input and predicts an RGB residual $\Delta I$ to yield the final refined image $\hat{I} = I_{\text{base}} + \Delta I$.
This design decouples compact 3D scene modeling from high-frequency appearance recovery, while preserving multi-view consistency: 
since both $I_{\text{base}}$ and $F_{\text{splat}}$ are rasterized from the same 3D Gaussians, the residual $\Delta I$ inherits its geometric structure from the underlying 3D representation rather than being hallucinated in 2D.

\section{Training}
We train \textbf{Anchor3DGS} on five public multi-view datasets: DL3DV~\citep{dl3dv}, ScanNet++~\citep{scannet++}, Hypersim~\citep{hypersim}, BlendedMVS~\citep{blendedmvs}, and Waymo~\citep{waymo}. These datasets cover synthetic and real-world captures, indoor and outdoor scenes, and object-scale to street-scale settings, which enhances the model's generalization ability to unseen scenarios.
For each training scene, we sample a set of context views as input. For datasets with sequentially ordered captures (\textit{e.g.,} DL3DV~\citep{dl3dv}), we adopt an \textit{index-based} sampler that selects two boundary frames at a chosen index gap, and samples additional context views from intermediate frame indices between them. For datasets without a sequential ordering (\textit{e.g.,} ScanNet++~\citep{scannet++}), we instead use a \textit{ranking-based} sampler that ranks all views by the camera-pose distance (rotation angle and translation) relative to a randomly chosen reference, and then selects context views from this distance-sorted list. In both cases, the extent covered by the selected context views (an index gap or a rank window) is randomized across iterations, producing a mixture of small- and wide-baseline configurations during training.

Our model is trained end-to-end by minimizing a combination of photometric and geometric losses. The photometric term consists of a pixel-wise MSE loss and a perceptual LPIPS~\citep{lpips} loss between the refined renderings $\hat{I}$ and the ground-truth images $I$. To regularize the geometry of the decoded Gaussians, we additionally apply a depth-consistency loss that encourages the depth rasterized from the Gaussians to agree with the depth maps predicted by the geometry backbone in Sec.~\ref{sec:geometry}. Therefore, the full training loss is:
\begin{equation}
\mathcal{L} = \lambda_{\text{mse}} \mathcal{L}_{\text{mse}} + \lambda_{\text{lpips}} \mathcal{L}_{\text{lpips}} + \lambda_{\text{depth}} \mathcal{L}_{\text{depth}}.
\end{equation}

\section{Experiments}

\begin{table}[t]
\centering
\caption{Quantitative comparison of the \textbf{sparse-view} novel view synthesis (3 and 6 input views) on VR-NeRF~\citep{vrnerf}, MipNeRF360~\citep{mipnerf360}, and DL3DV~\citep{dl3dv}. We report PSNR, SSIM, LPIPS, and the number of predicted Gaussian primitives (\#GS). \textbf{Bold} indicates the best.}
\vspace{4pt}
\label{tab:sparse}
\footnotesize
\setlength{\tabcolsep}{2.8pt}
\renewcommand{\arraystretch}{1.1}
\begin{tabular}{c|c|cccc|cccc|cccc}
\toprule
\multirow{2}{*}{Setting} & \multirow{2}{*}{Method} & \multicolumn{4}{c|}{VR-NeRF} & \multicolumn{4}{c|}{MipNeRF360} & \multicolumn{4}{c}{DL3DV} \\
\cline{3-14}
 & & PSNR & SSIM & LPIPS & \#GS
 & PSNR & SSIM & LPIPS & \#GS
 & PSNR & SSIM & LPIPS & \#GS \\
\midrule
\multirow{5}{*}{\rotatebox{90}{3 views}}
 & NoPoSplat~\citep{noposplat}        & 20.52 & 0.686 & 0.433 & 0.60M & 16.89 & 0.449 & 0.493 & 0.60M & 17.29 & 0.509 & 0.451 & 0.60M \\
 & FLARE~\citep{flare}                & 20.64 & 0.692 & 0.452 & 0.60M & 16.92 & 0.447 & 0.498 & 0.60M & 16.08 & 0.468 & 0.498 & 0.60M \\
 & DA3~\citep{da3}                    & 19.87 & 0.656 & 0.464 & 0.60M & 12.84 & 0.330 & 0.634 & 0.60M & 18.24 & 0.510 & 0.376 & 0.60M \\
 & AnySplat~\citep{anysplat}          & 21.31 & 0.729 & 0.340 & 0.49M & 15.65 & 0.624 & 0.331 & 0.38M & 17.74 & 0.625 & 0.308 & 0.51M \\
 & Ours                          & \textbf{23.69} & \textbf{0.764} & \textbf{0.320} & 0.26M
                                      & \textbf{21.36} & \textbf{0.678} & \textbf{0.307} & 0.26M
                                      & \textbf{21.22} & \textbf{0.683} & \textbf{0.280} & 0.26M \\
\midrule
\multirow{5}{*}{\rotatebox{90}{6 views}}
 & NoPoSplat~\citep{noposplat}        & 20.61 & 0.712 & 0.445 & 1.20M & 16.81 & 0.457 & 0.494 & 1.20M & 15.93 & 0.477 & 0.545 & 1.20M \\
 & FLARE~\citep{flare}                & 19.96 & 0.697 & 0.478 & 1.20M & 15.38 & 0.409 & 0.583 & 1.20M & 15.36 & 0.457 & 0.564 & 1.20M \\
 & DA3~\citep{da3}                    & 21.32 & 0.690 & 0.403 & 1.20M & 18.40 & 0.472 & 0.389 & 1.20M & 18.97 & 0.539 & 0.349 & 1.20M \\
 & AnySplat~\citep{anysplat}          & 22.45 & 0.758 & 0.321 & 1.03M & 19.01 & 0.613 & 0.283 & 0.70M & 18.42 & 0.613 & 0.316 & 0.98M \\
 & Ours                          & \textbf{24.03} & \textbf{0.781} & \textbf{0.312} & 0.26M
                                      & \textbf{22.39} & \textbf{0.663} & \textbf{0.278} & 0.26M
                                      & \textbf{20.70} & \textbf{0.649} & \textbf{0.303} & 0.26M \\
\bottomrule
\end{tabular}
\end{table}

\subsection{Experimental Setup}

\paragraph{Evaluation Datasets \& Baselines.}
Following AnySplat~\citep{anysplat}, we evaluate the novel view synthesis quality on VR-NeRF~\citep{vrnerf} and MipNeRF360~\citep{mipnerf360} datasets. We also test on a held-out subset of DL3DV~\citep{dl3dv} that is disjoint from training. Together, these benchmarks jointly cover object-centric, indoor, and outdoor scenes, providing a comprehensive evaluation.

We compare \textbf{Anchor3DGS} against recent pose-free feed-forward 3DGS methods: NoPoSplat~\citep{noposplat}, FLARE~\citep{flare}, Depth Anything 3 (DA3)~\citep{da3}, and AnySplat~\citep{anysplat}. Among them, NoPoSplat and FLARE are tailored for sparse-view inputs and do not scale to dense-view inputs, so we only include them in the sparse-view comparison (3 and 6 views). In the dense-view setting ($32$ views), we also report VGGT+3DGS as a per-scene optimization reference.
We compute PSNR, SSIM, and LPIPS between the rendered images and the ground-truth images. To further reflect the compactness of our representation, we also report the number of Gaussian primitives (\textit{\#GS}) for each scene.

\paragraph{Implementation Details.}
We set the anchor budget to $N=32{,}768$ and decode $K=8$ Gaussians per anchor, yielding roughly $0.26$M Gaussians per scene independent of input resolution and view count. The dimension of each anchor embedding $\mathbf{z}_n$ is $128$, and the dimension of per-Gaussian feature vector $\mathbf{h}$ (used for rendering refinement) is $16$.
We initialize the geometry backbone of \textbf{Anchor3DGS} with the pretrained VGGT~\citep{vggt} weights and fine-tune it jointly with the rest of the network using a $0.1\times$ smaller learning rate, while keeping the patch-embedding layer frozen. The model is trained on $8$ NVIDIA H20 GPUs for $30{,}000$ iterations (about $2$--$3$ days). We use the AdamW optimizer with a peak learning rate of $2 \times 10^{-4}$ and $1{,}000$ warmup steps. During training, the number of context views per scene is randomly sampled between $2$ and $16$, and the maximum input resolution is set to $448$ pixels on the longer side. At inference time, all images are resized to $448 \times 448$ resolution. For the training loss weights, we set $\lambda_{\text{mse}}=1.0$, $\lambda_{\text{lpips}}=0.05$, and $\lambda_{\text{depth}}=0.1$.

\subsection{Quantitative Results}
We first evaluate \textbf{Anchor3DGS} in the sparse-view setting (3 and 6 input views). As shown in Tab.~\ref{tab:sparse}, our method consistently achieves the best PSNR, SSIM, and LPIPS across all three benchmark datasets. The improvements are most pronounced on MipNeRF360 and DL3DV, where \textbf{Anchor3DGS} achieves PSNR gains of over $2$ dB compared to the best baseline under each setting.
These gains stem from two advantages of our anchor-based design: entropy-guided sampling concentrates capacity on visually complex regions, while cross-anchor self-attention propagates context to poorly-observed areas, mitigating the holes prevalent in pixel-aligned methods.
Notably, \textbf{Anchor3DGS} achieves these gains with only $0.26$M Gaussians per scene, showing that a compact anchor-based representation can yield both higher fidelity and greater efficiency in the sparse-view setting.

Furthermore, we test \textbf{Anchor3DGS} in the dense-view setting ($32$ input views), where the input already provides rich coverage of the scene.
As shown in Tab.~\ref{tab:dense}, our method provides comparable rendering quality against state-of-the-art pixel-aligned baselines, while requiring an order of magnitude fewer Gaussians and smaller storage footprint. Since the anchor number is decoupled from the input views, our \#GS stays at $0.26$M regardless of view count, whereas pixel-aligned methods produce several million Gaussians with no corresponding quality gain.
The VGGT+3DGS also provides a strong reference but represents a different paradigm—it requires minutes of optimization per scene, whereas \textbf{Anchor3DGS} directly predicts the 3DGS representation from the input views in a feed-forward pass within seconds. Despite this efficiency gap, our method still achieves on-par rendering quality.

\begin{table}[t]
\centering
\caption{\textbf{Dense-view} rendering results ($32$ input views) on VR-NeRF~\citep{vrnerf} and DL3DV~\citep{dl3dv}. We also report the storage size of the predicted Gaussians and the inference time on a single H800 GPU.}
\vspace{4pt}
\label{tab:dense}
\footnotesize
\setlength{\tabcolsep}{3.0pt}
\renewcommand{\arraystretch}{1.1}
\begin{tabular}{l|cccccc|cccccc}
\toprule
\multirow{2}{*}{Method} & \multicolumn{6}{c|}{VR-NeRF} & \multicolumn{6}{c}{DL3DV} \\
\cline{2-13}
 & PSNR & SSIM & LPIPS & \#GS & Storage & Time
 & PSNR & SSIM & LPIPS & \#GS & Storage & Time \\
\midrule
VGGT+3DGS~\citep{3DGS}             & 23.08 & 0.762 & 0.301 & 0.44M & 109MB & 6min
                                   & 19.34 & \textbf{0.627} & \textbf{0.312} & 1.16M & 288MB & 8min \\
\midrule
DA3~\citep{da3}                    & 22.38 & 0.751 & 0.350 & 6.42M & 1.03GB & 1.9s
                                   & 19.38 & 0.544 & 0.341 & 6.42M & 1.03GB & 2.0s \\
AnySplat~\citep{anysplat}          & 24.05 & \textbf{0.819} & 0.309 & 4.75M & 779MB & 2.8s
                                   & 18.76 & 0.565 & 0.371 & 3.98M & 653MB & 2.8s \\
Ours                               & \textbf{24.56} & 0.815 & \textbf{0.297} & 0.26M & 35MB & 2.9s
                                   & \textbf{19.52} & 0.593 & 0.364 & 0.26M & 35MB & 3.1s \\
\bottomrule
\end{tabular}
\end{table}

\begin{figure}[t]
    \centering
    \includegraphics[width=0.98\linewidth]{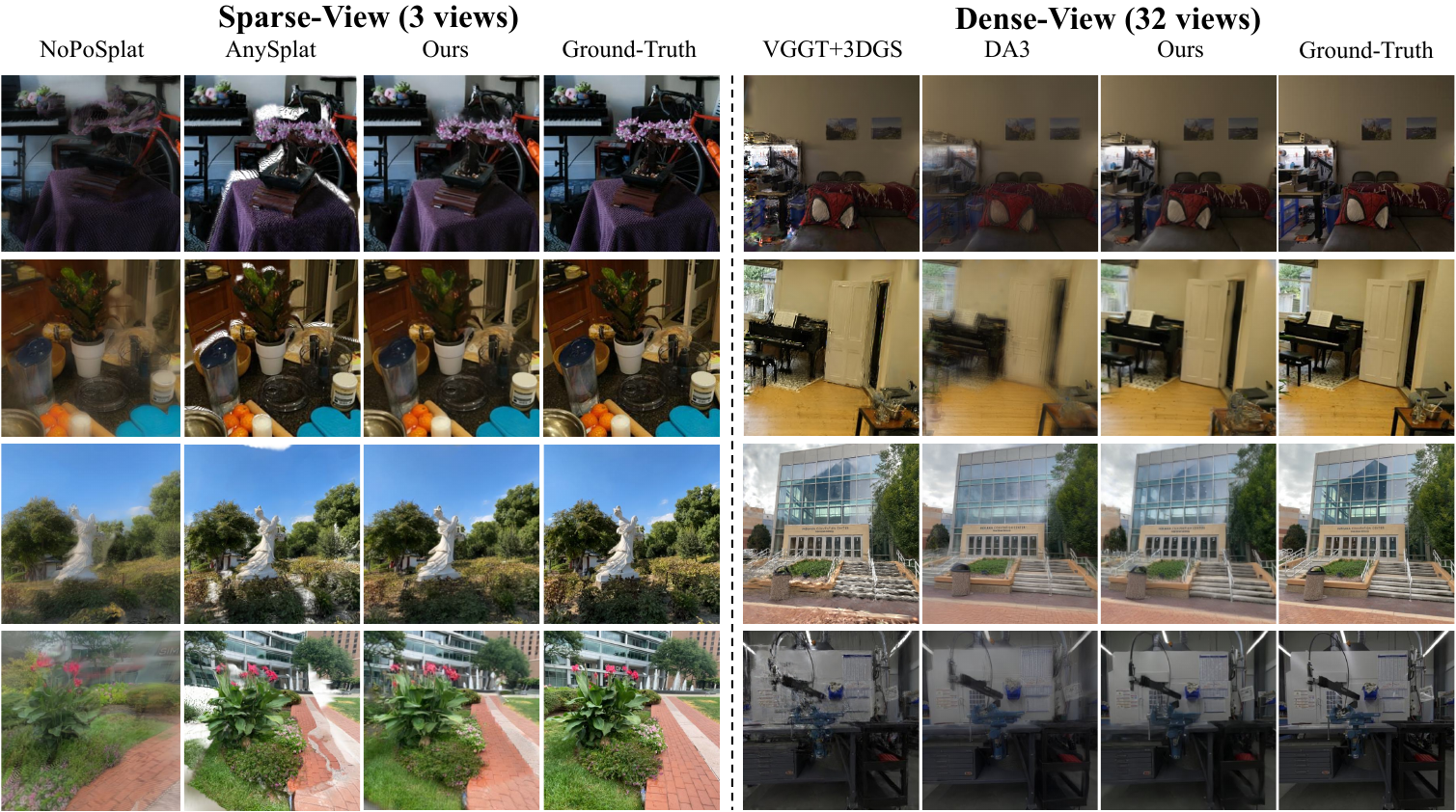}
    \caption{Qualitative comparison of novel view synthesis. Our method recovers sharper textures and finer details while mitigating artifacts in poorly observed regions.}
    \label{fig:qual_comparison}
\end{figure}

\subsection{Qualitative Results}

Fig.~\ref{fig:qual_comparison} presents qualitative comparisons under both sparse-view and dense-view settings. In the sparse-view setting, the representative method NoPoSplat~\citep{noposplat} tends to produce overly blurry renderings, while AnySplat~\citep{anysplat} frequently leaves holes in poorly observed regions. Our method alleviates both issues, yielding coherent structure and visually appealing results across different scenes.
In the dense-view setting, our renderings remain on par with the state-of-the-art baselines while using an order of magnitude fewer Gaussian primitives. Our method recovers sharp textures and fine details, and avoids the fragmented, needle-like artifacts commonly exhibited by optimization-based 3DGS pipelines.
More visualization results are provided in the Appendix.

\subsection{Ablation Study}

\begin{table}[t]
\centering
\caption{Ablation study results on VR-NeRF~\citep{vrnerf} dataset. \textbf{Bold} indicates the best.}
\vspace{4pt}
\label{tab:ablation}
\small
\setlength{\tabcolsep}{9.5pt}
\renewcommand{\arraystretch}{1.1}
\begin{tabular}{l|ccc|ccc}
\toprule
\multirow{2}{*}{Configuration} & \multicolumn{3}{c|}{3 views} & \multicolumn{3}{c}{32 views} \\
\cline{2-7}
 & PSNR $\uparrow$ & SSIM $\uparrow$ & LPIPS $\downarrow$
 & PSNR $\uparrow$ & SSIM $\uparrow$ & LPIPS $\downarrow$ \\
\midrule
w/o enhancement          & 22.50 & 0.754 & 0.345 & 24.13 & 0.807 & 0.324 \\
w/o self-attention         & 23.03 & 0.750 & 0.323 & 24.38 & 0.808 & 0.303 \\
w/o occ-aggregation        & 23.28 & 0.756 & 0.330 & 24.29 & 0.813 & 0.305 \\
w/o entropy-sample         & 22.16 & 0.733 & 0.355 & 24.17 & 0.804 & 0.318 \\
\midrule
Full model (Ours)          & \textbf{23.69} & \textbf{0.764} & \textbf{0.320} & \textbf{24.56} & \textbf{0.815} & \textbf{0.297} \\
\bottomrule
\end{tabular}
\end{table}

\begin{figure}[t]
    \centering
    \includegraphics[width=0.99\linewidth]{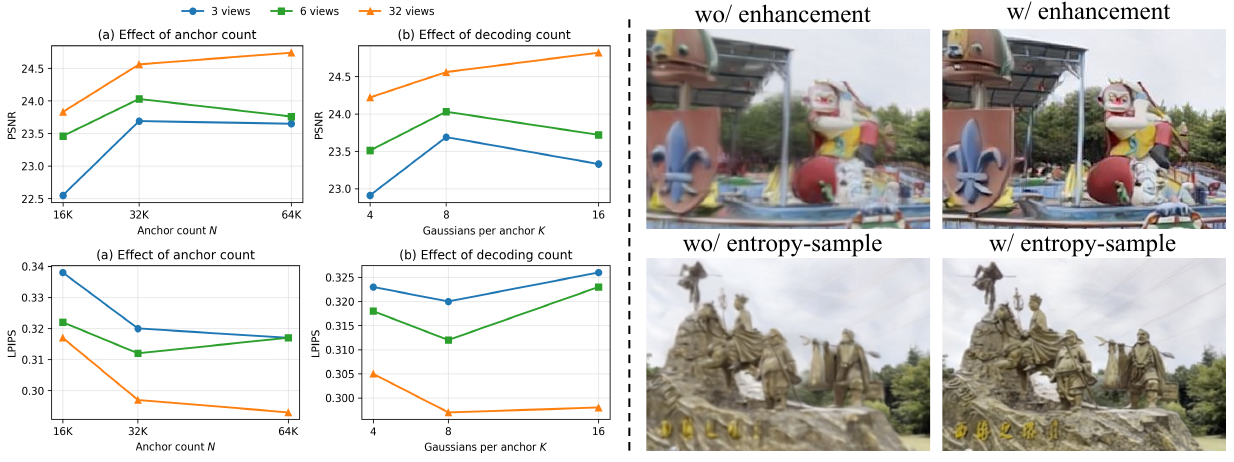}
    \caption{Rendering metrics under varying anchor count $N$ and decoded Gaussians $K$ (\textit{left}), and qualitative comparisons of removing the enhancement module and entropy-guided sampling (\textit{right}).}
    \label{fig:ablation}
\end{figure}

We conduct the ablation study to validate each key design of \textbf{Anchor3DGS}. Tab.~\ref{tab:ablation} reports results on VR-NeRF~\citep{vrnerf} under both sparse-view ($3$ views) and dense-view ($32$ views) settings. Removing any component causes a quality drop, confirming that all four designs contribute to the final performance. 
\textit{(i) w/o enhancement} removes the enhancement UNet for refinement, which leads to consistent degradation as the splatted RGB alone struggles to recover fine high-frequency details. \textit{(ii) w/o self-attention} disables cross-anchor interaction so each anchor is encoded independently, and \textit{(iii) w/o occ-aggregation} replaces occlusion-aware multi-view aggregation with plain mean pooling. Both of them impair the encoded anchor features and thus degrade rendering performance. \textit{(iv) w/o entropy-sample} replaces entropy-guided sampling with uniform grid subsampling, which causes noticeable PSNR drop ($-1.53$ dB at $3$ views). This indicates that our strategy allocates the limited anchor budget more reasonably.
We further examine how performance varies with anchor count $N$ and decoded Gaussians per anchor $K$ in Fig.~\ref{fig:ablation}. The results confirm that our default setting ($N=32{,}768$ and $K{=}8$) effectively balances rendering quality and Gaussian compactness.

\section{Conclusion}
We present \textbf{Anchor3DGS}, a pose-free feed-forward 3DGS framework that replaces the prevailing pixel-aligned paradigm with a compact anchor-based paradigm. By placing a fixed number of 3D anchors on a geometry-aware scaffold, our method decouples the Gaussian count from the input resolution and view number, and instead allocates representation capacity according to scene content. Three key designs—entropy-guided anchor sampling, occlusion-aware multi-view aggregation, and feature-based rendering refinement—together enable high-fidelity novel view synthesis. Extensive experiments across multiple datasets demonstrate that \textbf{Anchor3DGS} achieves state-of-the-art rendering quality in sparse-view settings. In dense-view settings, \textbf{Anchor3DGS} attains comparable quality while using an order of magnitude fewer Gaussians and lower storage cost, validating the effectiveness and efficiency of our compact anchor-based representation. Beyond view synthesis, such a compact anchor-based representation naturally serves as a global geometric memory~\cite{machine-memory}, offering a promising substrate for future video world models.

Despite the advancements, \textbf{Anchor3DGS} still has limitations. The performance of our method relies on the prediction quality of the geometry backbone. When the backbone produces inaccurate poses or noisy depth maps, the resulting anchor placement may degrade and harm the final rendering. In addition, since the anchor budget is fixed regardless of scene scale, very large or unbounded scenes may benefit from a scene-adaptive budget, which we leave as future work.

\begin{ack}
This work was supported by the National Natural Science Foundation of China (62332019, 62461160309) and NSFC-RGC Joint Research Scheme N\_HKU705/24.
\end{ack}


\bibliographystyle{plainnat}
\bibliography{references}


\newpage
\appendix
\begin{center}
    {\Large\bfseries Appendix}
\end{center}
\vspace{1em}

\setcounter{figure}{0}
\setcounter{table}{0}
\renewcommand{\thefigure}{A\arabic{figure}}
\renewcommand{\thetable}{A\arabic{table}}

\section{More Implementation Details}
\label{app:impl}

\paragraph{Model Architecture.}
Our geometry backbone is a VGGT-based~\citep{vggt} model with a ViT architecture. The backbone takes $V$ uncalibrated images as input and produces multi-scale tokens, from which we take the intermediate features at layers $\{4, 11, 17, 23\}$ and fuse them through a DPT-style head into per-view feature maps $\mathbf{F}_v\!\in\!\mathbb{R}^{128\times H\times W}$ used for anchor feature aggregation.
For anchor encoding, we first apply Fourier positional encoding with $L{=}6$ frequencies (yielding $3+6\!\times\!3\!\times\!2{=}39$ dimensions) on the anchor coordinates and concatenate $\log d_n$ as an additional density signal.
After multi-view aggregation, each anchor feature is refined by $4$ self-attention blocks, where each block uses RMSNorm pre-normalization, multi-head self-attention with $\text{head\_dim}{=}128$ ($4$ heads), and an MLP layer with expansion ratio $4$.
The feature enhancement module is a 3-level UNet with channel widths $\{64, 128, 256\}$ from shallow to deep. Downsampling uses $3\!\times\!3$ convolutions with stride $2$, upsampling uses $4\!\times\!4$ transposed convolutions followed by a $3\!\times\!3$ convolution, and skip connections concatenate encoder and decoder features at each level. A final $1\!\times\!1$ convolution produces the $3$-channel RGB residual $\Delta I$. The weights and biases of the final convolution are zero-initialized, so the network behaves as an identity mapping at the start of training.

\paragraph{Datasets.}
We train on five public multi-view datasets covering diverse scene types: DL3DV~\citep{dl3dv} (real-world indoor and outdoor scenes), ScanNet++~\citep{scannet++} (real-world indoor scenes), Hypersim~\citep{hypersim} (synthetic indoor scenes), BlendedMVS~\citep{blendedmvs} (object-centric and aerial scenes), and Waymo~\citep{waymo} (outdoor driving scenes). These datasets are mixed with sampling probabilities $0.4, 0.2, 0.2, 0.1, 0.1$ respectively.
All training images are resized so that the longer side equals $448$ pixels, with intrinsics rescaled accordingly. For evaluation, following AnySplat~\citep{anysplat}, we report results on three benchmarks: VR-NeRF~\citep{vrnerf} ($4$ scenes), MipNeRF360~\citep{mipnerf360} ($4$ scenes), and DL3DV~\citep{dl3dv} ($10$ held-out scenes).

\paragraph{Testing Protocol.}
Because our task takes uncalibrated images as input, evaluation must rely on camera poses predicted by the model itself. We follow the testing protocol of AnySplat~\citep{anysplat}.
Given a test scene with context views and held-out target views, we run the model twice: \textit{(i)} On context views alone, we obtain the predicted context poses $\mathcal{C}^{(1)}$ and the scene's 3D Gaussians $\mathcal{G}$. \textit{(ii)} On the union of context and target views, we obtain a second set of poses $\mathcal{C}^{(2)}$ that includes the target views. Since the two passes can produce slightly different context poses, we apply a Procrustes alignment between the context poses in $\mathcal{C}^{(1)}$ and $\mathcal{C}^{(2)}$, and use the resulting transform to map the target poses in $\mathcal{C}^{(2)}$ into the coordinate frame of $\mathcal{G}$. We then render $\mathcal{G}$ at the aligned target poses and compute metrics against ground-truth images.
For NoPoSplat~\citep{noposplat}, which does not predict input camera poses, we follow the original protocol and recover target camera poses via a $200$-step test-time optimization that minimizes the MSE between the rendered image and the ground-truth image.

\section{Additional Experimental Results}
\label{app:exp}

\subsection{Camera Pose Estimation}
Since our geometry backbone also recovers camera parameters, we also evaluate pose estimation on the Co3Dv2~\citep{co3d} dataset. Following~\citep{vggt, anysplat}, we adopt the AUC metric that computes relative pose accuracy over a range of angular thresholds, and set thresholds as $10$, $20$, and $30$ degrees.
As shown in Tab.~\ref{tab:pose}, our geometry backbone outperforms both AnySplat~\citep{anysplat} and the original VGGT~\citep{vggt} across all AUC thresholds. This suggests that jointly training the geometry backbone with our anchor-based 3DGS pipeline does not compromise the underlying pose estimation ability.

\begin{table}[ht]
\centering
\caption{Camera pose estimation results on Co3Dv2~\citep{co3d} dataset. We report AUC at angular thresholds of $10$, $20$, and $30$ degrees. \textbf{Bold} indicates the best.}
\vspace{4pt}
\label{tab:pose}
\small
\setlength{\tabcolsep}{15pt}
\renewcommand{\arraystretch}{1.2}
\begin{tabular}{l|ccc}
\toprule
Method & AUC@$10$ $\uparrow$ & AUC@$20$ $\uparrow$ & AUC@$30$ $\uparrow$ \\
\midrule
VGGT~\citep{vggt}                  & 50.4 & 67.2 & 74.9 \\
AnySplat~\citep{anysplat}          & 53.1 & 73.1 & 80.8 \\
Ours                               & \textbf{54.8} & \textbf{74.0} & \textbf{81.5} \\
\bottomrule
\end{tabular}
\end{table}

\subsection{Novel-View Synthesis with Denser Views}
To further examine how our anchor-based representation scales with denser inputs, we extend the dense-view evaluation to $64$ context views on VR-NeRF~\citep{vrnerf} and DL3DV~\citep{dl3dv} datasets. Results are reported in Tab.~\ref{tab:64view}. Consistent with the $32$-view setting, our method remains competitive with or better than prior pixel-aligned feed-forward baselines (\textit{i.e.,} DA3~\citep{da3}, AnySplat~\citep{anysplat}) in rendering metrics while using roughly one to two orders of magnitude fewer Gaussian primitives.

\begin{table}[ht]
\centering
\caption{Novel-view synthesis with $64$ context views on VR-NeRF~\citep{vrnerf} and DL3DV~\citep{dl3dv}. We also report the number (\#GS) and storage size of the predicted Gaussians.}
\vspace{4pt}
\label{tab:64view}
\small
\setlength{\tabcolsep}{2.9pt}
\renewcommand{\arraystretch}{1.2}
\begin{tabular}{l|ccccc|ccccc}
\toprule
\multirow{2}{*}{Method} & \multicolumn{5}{c|}{VR-NeRF} & \multicolumn{5}{c}{DL3DV} \\
\cline{2-11}
 & PSNR & SSIM & LPIPS & \#GS & Storage
 & PSNR & SSIM & LPIPS & \#GS & Storage \\
\midrule
VGGT+3DGS~\citep{3DGS}             & 23.74 & 0.780 & 0.289 & 0.46M & 114MB
                                   & 20.69 & 0.667 & 0.270 & 1.39M & 345MB \\
\midrule
DA3~\citep{da3}                    & 21.91 & 0.725 & 0.365 & 12.8M & 2.10GB
                                   & 19.64 & 0.564 & 0.328 & 12.8M & 2.10GB \\
AnySplat~\citep{anysplat}          & 23.63 & 0.797 & 0.318 & 8.97M & 1.47GB
                                   & 18.71 & 0.567 & 0.369 & 7.43M & 1.22GB \\
Ours                               & 24.01 & 0.799 & 0.311 & 0.26M & 35MB
                                   & 19.33 & 0.585 & 0.367 & 0.26M & 35MB \\
\bottomrule
\end{tabular}
\end{table}

\subsection{Rendering Speed Analysis}
We further investigate whether our feature-enhancement UNet can become a rendering-time bottleneck.
To quantify its impact, we compare the per-frame rendering speed of our method against two pixel-aligned feed-forward baselines on VR-NeRF~\citep{vrnerf} dataset. For each view configuration, we run the model to obtain the predicted Gaussians, then measure the rendering speed over $100$ rendering passes on a single H20 GPU. The averaged FPS is reported in Tab.~\ref{tab:fps}.

\begin{table}[ht]
\centering
\caption{Rendering speed analysis on VR-NeRF~\citep{vrnerf} dataset. \#GS denotes the number of predicted Gaussians; FPS is the per-frame 3DGS rendering speed averaged over $100$ iterations.}
\vspace{4pt}
\label{tab:fps}
\small
\setlength{\tabcolsep}{10pt}
\renewcommand{\arraystretch}{1.2}
\begin{tabular}{l|cc|cc|cc}
\toprule
\multirow{2}{*}{Method} & \multicolumn{2}{c|}{6 views} & \multicolumn{2}{c|}{32 views} & \multicolumn{2}{c}{64 views} \\
\cline{2-7}
 & \#GS & FPS $\uparrow$ & \#GS & FPS $\uparrow$ & \#GS & FPS $\uparrow$ \\
\midrule
DA3~\citep{da3}            & 1.20M & 361.1 & 6.42M & 146.3 & 12.8M & 85.9  \\
AnySplat~\citep{anysplat}  & 1.03M & 375.4 & 4.75M & 176.7 & 8.97M & 120.5 \\
Ours                       & 0.26M & 179.4 & 0.26M & 190.6 & 0.26M & 187.4 \\
\bottomrule
\end{tabular}
\end{table}

The enhancement UNet is lightweight, so our rendering speed still notably exceeds real-time requirements. Because our anchor-based representation produces a constant $\sim$0.26M Gaussians regardless of the number of input views, our rendering speed stays stable across all configurations. In contrast, pixel-aligned baselines produce \#GS proportional to input views, and their FPS drops sharply from $\sim$360 (6 views) to under 86 (64 views). In particular, our FPS is $1.6\times$ faster than AnySplat and $2.2\times$ faster than DA3 at 64 views, and the gap can keep widening as the view count increases.

\subsection{Multi-View Consistency Analysis}
To verify the impact of the enhancement 2D UNet on multi-view consistency, we conduct a dedicated experiment. On an interpolated novel-view (virtual camera) trajectory, we measure the adjacent-view, depth-based reprojection warping error (RMSE) in Tab.~\ref{tab:consistency}, masked to co-visible and geometrically consistent pixels. The variant without the UNet and the UNet-refined render share identical geometry backbone and anchor-based decoder, so their difference isolates the UNet.

\begin{table}[ht]
\centering
\caption{Multi-view consistency analysis on VR-NeRF~\cite{vrnerf} dataset. We report the RMSE metric.}
\vspace{4pt}
\label{tab:consistency}
\small
\setlength{\tabcolsep}{20pt}
\renewcommand{\arraystretch}{1.2}
\begin{tabular}{c|cc}
\toprule
Warping RMSE $\downarrow$ (traj) & with UNet & without UNet (reference) \\
\midrule
6 input views                 & 0.01187 & 0.01018 \\
32 input views                & 0.00974 & 0.00970 \\
\bottomrule
\end{tabular}
\end{table}

The UNet clearly improves rendering quality, while the increase in warping RMSE over the variant without the UNet (consistent reference) is marginal, indicating that it introduces minimal multi-view inconsistency. The reason is structural: the feature map fed to the UNet is rasterized from the same 3D Gaussians with the same alpha-compositing weights as the base image, so the predicted residual inherits the 3D geometric structure of the underlying representation rather than being hallucinated in 2D image space. Furthermore, the enhancement 2D UNet is lightweight, which inherently limits its capacity to introduce multi-view artifacts.

\subsection{More Ablation Results}
\begin{figure}[ht]
    \centering
    \includegraphics[width=\linewidth]{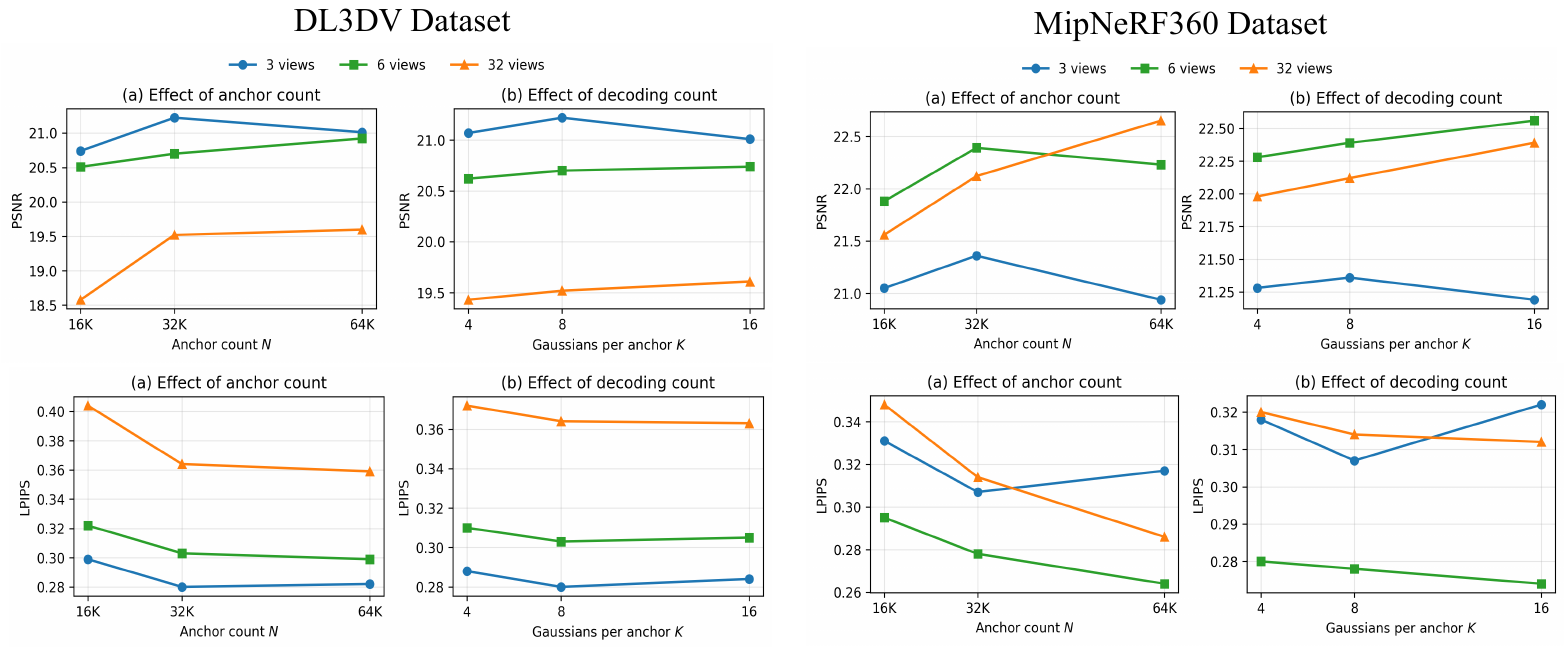}
    \caption{Rendering metrics under varying anchor count $N$ and decoded Gaussians per anchor $K$ on DL3DV~\citep{dl3dv} and MipNeRF360~\citep{mipnerf360} datasets.}
    \label{fig:appendix_ablation}
\end{figure}

We provide additional ablation studies besides the analysis in the main paper. Tab.~\ref{tab:ablation_dl3dv} reports the component ablation on DL3DV~\citep{dl3dv} under $3$, $6$, and $32$ input views, and Fig.~\ref{fig:appendix_ablation} shows the rendering metrics under varying anchor count $N$ and decoded Gaussians per anchor $K$ on DL3DV~\citep{dl3dv} and MipNeRF360~\citep{mipnerf360}. The conclusions are consistent with the main paper: removing any single component (\textit{UNet enhancement}, \textit{anchor attention}, \textit{occlusion-aware aggregation}, or \textit{entropy-sample}) consistently degrades rendering quality, and our default setting ($N{=}32{,}768$ and $K{=}8$) provides a favorable balance between rendering performance and Gaussian compactness.

\begin{table}[ht]
\centering
\caption{Ablation study on DL3DV~\citep{dl3dv} dataset under $3$, $6$, and $32$ input views.}
\vspace{4pt}
\label{tab:ablation_dl3dv}
\small
\setlength{\tabcolsep}{4pt}
\renewcommand{\arraystretch}{1.2}
\begin{tabular}{l|ccc|ccc|ccc}
\toprule
\multirow{2}{*}{Configuration}
 & \multicolumn{3}{c|}{3 views}
 & \multicolumn{3}{c|}{6 views}
 & \multicolumn{3}{c}{32 views} \\
\cline{2-10}
 & PSNR & SSIM & LPIPS
 & PSNR & SSIM & LPIPS
 & PSNR & SSIM & LPIPS \\
\midrule
w/o enhancement                       & 18.81 & 0.647 & 0.334 & 19.62 & 0.622 & 0.358 & 18.66 & 0.564 & 0.424 \\
w/o self-attention                    & 20.88 & 0.673 & 0.292 & 20.81 & 0.648 & 0.312 & 19.43 & 0.588 & 0.375 \\
w/o occ-aggregation                   & 21.00 & 0.675 & 0.295 & \textbf{20.72} & 0.642 & 0.317 & 19.03 & 0.576 & 0.390 \\
w/o entropy-sample (uniform)          & 20.93 & 0.669 & 0.290 & 20.67 & 0.634 & 0.309 & 19.21 & 0.580 & 0.373 \\
w/o entropy-sample (voxel)            & 20.32 & 0.630 & 0.345 & 19.73 & 0.594 & 0.376 & 18.34 & 0.537 & 0.446 \\
\midrule
Full model (Ours)                     & \textbf{21.22} & \textbf{0.683} & \textbf{0.280} & 20.70 & \textbf{0.649} & \textbf{0.303} & \textbf{19.52} & \textbf{0.593} & \textbf{0.364} \\
\bottomrule
\end{tabular}
\end{table}

Furthermore, we investigate whether the improvement stems from the anchor representation itself or from the fine-tuned geometry backbone. We keep the identical fine-tuned VGGT backbone, and replace only the prediction head: the anchor head is swapped for a standard pixel-aligned decoder (one Gaussian per pixel, as in AnySplat~\cite{anysplat}). We report the average metrics in Tab.~\ref{tab:ablate_backbone_anchor}.

\begin{table}[t]
\centering
\caption{Ablation study on the anchor representation. We compare our anchor-based head with a standard pixel-aligned decoder under the identical fine-tuned VGGT backbone. }
\vspace{4pt}
\label{tab:ablate_backbone_anchor}
\footnotesize
\setlength{\tabcolsep}{4pt}
\renewcommand{\arraystretch}{1.3}
\begin{tabular}{c|c|ccc|ccc|ccc}
\toprule
\multirow{2}{*}{Dataset} & \multirow{2}{*}{Setting} & \multicolumn{3}{c|}{3 views} & \multicolumn{3}{c|}{6 views} & \multicolumn{3}{c}{32 views} \\
\cline{3-11}
 & & PSNR & SSIM & LPIPS
 & PSNR & SSIM & LPIPS
 & PSNR & SSIM & LPIPS \\
\midrule
\multirow{3}{*}{\rotatebox{90}{VR-NeRF}}
 & Backbone + Pixel Decoder       & 21.59 & 0.736 & 0.339 & 22.48 & 0.760 & 0.319 & 24.24 & 0.818 & 0.304 \\
 & AnySplat~\citep{anysplat}      & 21.31 & 0.729 & 0.340 & 22.45 & 0.758 & 0.321 & 24.05 & 0.819 & 0.309 \\
 & Anchor3DGS (Ours)              & 23.69 & 0.764 & 0.320 & 24.03 & 0.781 & 0.312 & 24.56 & 0.815 & 0.297 \\
\midrule
\multirow{3}{*}{\rotatebox{90}{DL3DV}}
 & Backbone + Pixel Decoder       & 17.89 & 0.632 & 0.305 & 18.81 & 0.620 & 0.312 & 19.02 & 0.576 & 0.369 \\
 & AnySplat~\citep{anysplat}      & 17.74 & 0.625 & 0.308 & 18.42 & 0.613 & 0.316 & 18.76 & 0.565 & 0.371 \\
 & Anchor3DGS (Ours)              & 21.22 & 0.683 & 0.280 & 20.70 & 0.649 & 0.303 & 19.52 & 0.593 & 0.364 \\
\bottomrule
\end{tabular}
\end{table}

The results isolate two effects. (i) The backbone contribution — comparing our backbone + pixel decoder against AnySplat~\cite{anysplat} — is small: +0.28 / +0.03 / +0.19 dB on VR-NeRF~\cite{vrnerf} dataset and +0.15 / +0.39 / +0.26 dB on DL3DV~\cite{dl3dv} dataset at 3 / 6 / 32 views, confirming that fine-tuning the backbone alone does not explain our rendering improvement. (ii) The anchor contribution — comparing Anchor3DGS against our backbone + pixel decoder — is substantial: +2.10 dB (VR-NeRF 3 views) and +3.33 dB (DL3DV 3 views), narrowing to +0.32 / +0.50 dB at 32 views. The performance gain therefore primarily comes from the anchor-based representation rather than the fine-tuned backbone. Unlike pixel-aligned decoders that predict one Gaussian per pixel and leave holes in poorly-observed regions, our anchors aggregate multi-view context through attention mechanism and concentrate capacity in high-entropy regions via entropy-guided sampling.

\section{Broader Social Impacts}
\textbf{Anchor3DGS} is a general-purpose feed-forward 3D reconstruction and novel view synthesis method, and we expect its social impact to be generally positive. By predicting a compact set of 3D Gaussians from uncalibrated images in a single forward pass, our method substantially reduces the per-scene optimization cost of conventional 3DGS pipelines. This efficiency lowers the computational and memory footprint of 3D content creation, and also makes high-quality 3D reconstruction more accessible to downstream applications such as virtual and augmented reality, robotics, autonomous driving simulation, and immersive telepresence.
On the other hand, like other photorealistic novel view synthesis techniques, our method could be misused. Realistic renderings of reconstructed scenes might be repurposed to fabricate misleading visual content, and unconstrained reconstruction from publicly available imagery raises potential privacy concerns regarding individuals or private spaces captured in the input views.
These risks can be mitigated through standard practices, including responsible data curation, watermarking or provenance tracking of synthesized content, and obtaining proper consent before reconstructing scenes that contain private property.

\section{More Visualizations}
\begin{figure}[ht]
    \centering
    \includegraphics[width=0.89\linewidth]{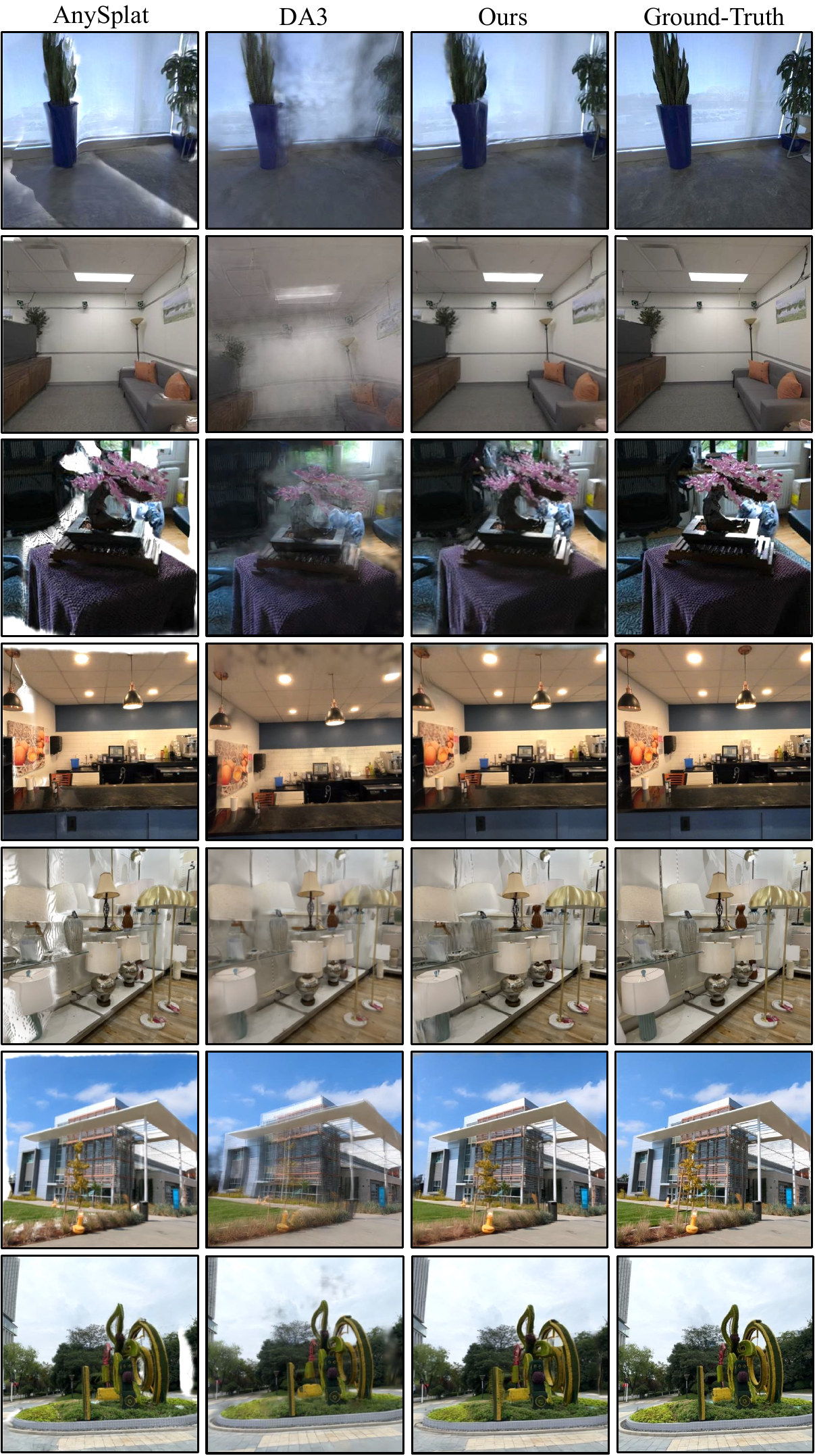}
    \caption{More qualitative comparisons against existing feed-forward methods.}
    \label{fig:appendix_vis1}
\end{figure}

We provide more visual comparisons with baseline methods in Fig.~\ref{fig:appendix_vis1}. Despite its compact representation, our method is still able to generate high-quality renderings with sharp textures and fine details. Please refer to our supplementary video for more novel view rendering results of our method.

\clearpage

\end{document}